\documentclass[journal]{IEEEtai}

\usepackage[colorlinks,urlcolor=blue,linkcolor=blue,citecolor=blue]{hyperref}

\usepackage{color,array}

\usepackage{graphicx}

\usepackage{amsmath}
\usepackage{amsfonts}
\usepackage{amssymb}

\usepackage{float}

\usepackage{algorithm}
\usepackage{algorithmic}

\begin{document}

\title{Complex Facial Expression Recognition Using Deep Knowledge Distillation of Basic Features}

\author{Angus Maiden and Bahareh Nakisa
\thanks{A. Maiden is with the School of Information Technology, Deakin University, Geelong, VIC 3220 Australia.}
\thanks{B. Nakisa is with the School of Information Technology, Deakin University, Geelong, VIC 3220 Australia.}}

\markboth{arXiv preprint}
{Angus Maiden and Bahareh Nakisa: Complex Facial Expression Recognition Using Deep Knowledge Distillation of Basic Features}

\maketitle

\begin{abstract}
Complex emotion recognition is a cognitive task that has so far eluded the same excellent performance of other tasks that are at or above the level of human cognition. Emotion recognition through facial expressions is particularly difficult due to the complexity of emotions expressed by the human face. For a machine to approach the same level of performance in complex facial expression recognition as a human, it may need to synthesise knowledge and understand new concepts in real-time, as humans do. Humans are able to learn new concepts using only few examples by distilling important information from memories. Inspired by human cognition and learning, we propose a novel continual learning method for complex facial expression recognition that can accurately recognise new compound expression classes using few training samples, by building on and retaining its knowledge of basic expression classes. In this work, we also use GradCAM visualisations to demonstrate the relationship between basic and compound facial expressions. Our method leverages this relationship through knowledge distillation and a novel Predictive Sorting Memory Replay, to achieve the current state-of-the-art in continual learning for complex facial expression recognition, with 74.28\% Overall Accuracy on new classes. We also demonstrate that using continual learning for complex facial expression recognition achieves far better performance than non-continual learning methods, improving on state-of-the-art non-continual learning methods by 13.95\%. Our work is also the first to apply few-shot learning to complex facial expression recognition, achieving the state-of-the-art with 100\% accuracy using only a single training sample per class.
\end{abstract}

\begin{IEEEImpStatement}
Facial expressions are one of the most powerful signals for humans to convey emotional states, comprising over 55\% of our emotional communication. By developing AI systems that can accurately recognise facial expressions at human performance level, they could be trusted to assist with functions and services that demand emotional communication, such as in healthcare and customer service. We propose a novel method of complex facial expression recognition using continual learning and few-shot learning, inspired by human cognition and learning. The method uses knowledge distillation of basic expressions and a novel Predictive Sorting Memory Replay to reduce the catastrophic forgetting associated with continual learning, achieving state-of-the-art performance compared with other methods. Our method demonstrates the increased performance of neural networks when recognising complex concepts by retaining and distilling the knowledge of basic concepts. This will pave the way for further research and applications in other domains using our method.
\end{IEEEImpStatement}


\section{Introduction}
\label{sec:1}

We are now entering the fourth industrial age, where artificial intelligence (AI) plays a crucial role in many of our activities and endeavours. By equalling or outperforming humans in cognitive tasks such as image recognition and natural language processing, and by completing physical and processing tasks with far greater precision and speed, AI can increasingly be used to assist with complex functions and services that were once the exclusive domain of humans, such as automobile driving, medical diagnosis, and customer service operations. However, many of these services rely not only on technical accuracy and precision, but on the very human elements of communication, empathy and compassion.

Communication between humans is fundamental to our ability to learn, work and build structures and societies, allowing us to survive, adapt, progress and prosper. According to Darwin \cite{ref:1}, facial expressions are one of the most powerful signals for humans to convey emotional states and intentions, with over 55\% of emotional communication being conveyed through facial expressions \cite{ref:2}. \textit{Facial expression recognition (FER)} is thus a crucial factor in our ability to operate in more complex and nuanced roles such as healthcare and customer service. By developing AI systems that can accurately recognise human emotional states at or above the level of human performance, they could be trusted to assist with these more complex functions in order to enhance these services.

In order for an AI system to recognise complex facial expressions at or above the level of human cognition, it may need to learn in a similar way to humans, learning new concepts from only a few examples by synthesising those concepts with existing knowledge. \textit{Continual learning} and \textit{few-shot learning} are two approaches to machine learning that are inspired by human cognition and learning patterns. Continual learning is an approach for incrementally learning new classes using a model that was previously trained on other classes. It uses techniques such as knowledge distillation and memory replay to retain the knowledge of known classes, such that it performs well in recognising both old and new classes. Few-shot learning involves training machine learning models using only a few examples. High-performing few-shot learning methods use novel feature extraction and data augmentation techniques to achieve high recognition accuracy using fewer training samples. Applying these methods of continual learning and few-shot learning to complex FER could lead to human-like performance on this challenging task.

This work investigates the hypothesis that by retaining the knowledge of basic facial expression features, a machine learning model will achieve better performance when learning new complex facial expression labels which share those features. The work is comprised of three phases. In the initial \textit{Basic FER Phase}, we design and build a neural network model that can achieve high classification accuracy on six basic expression classes. In the \textit{Continual Learning Phase}, new complex expression classes are iteratively learned by using knowledge distillation of basic expressions as well as a unique \textit{Predictive Sorting Memory Replay}. In the \textit{Few-shot Learning Phase}, we demonstrate how new complex expression classes can be learned to high accuracy with a very small number of training samples using knowledge distillation of basic expressions.

The main contributions of this work are:
\begin{itemize}
\item We propose a novel method of complex facial expression recognition using continual learning and few-shot learning. The method uses data augmentation, knowledge distillation of basic expressions, and a novel Predictive Sorting Memory Replay to reduce catastrophic forgetting and improve performance using few training examples.
\item We demonstrate that using continual learning for complex facial expression recognition achieves far better performance than non-continual learning methods, improving on the state-of-the-art in non-continual learning methods by 13.95\%.
\item We achieve the current state-of-the-art in continual learning for complex facial expression recognition, with 74.28\% Overall Accuracy on new classes (an improvement of 0.67\%) in a comprehensive analysis using experiments that compare with other continual learning methods.
\item Our method is the first to apply few-shot learning to complex facial expression recognition to the best of our knowledge, achieving the state-of-the-art with 100\% accuracy using a single training sample for each expression class.
\end{itemize}

\section{Literature Review}
\label{sec:2}

\subsection{Background and Scope}
\label{sec:2.1}

The majority of prior research on FER uses a categorical expression labelling system based on \cite{ref:3}, who defined six basic emotions expressed on the human face: \textit{anger}, \textit{disgust}, \textit{fear}, \textit{happiness}, \textit{sadness}, and \textit{surprise}, with \textit{contempt} later added by \cite{ref:4}. These basic expressions are apparently conveyed and perceived similarly across different cultures \cite{ref:5}. High recognition performance in recent years indicates that FER using basic expression labels is essentially solved (see Table \ref{tab:1}).

\begin{table}[!h]
\centering
\begin{tabular}{|p{0.5\columnwidth}|l|l|l|}
\hline 
\textbf{Method} & \textbf{Dataset} & \textbf{Classes} & \textbf{Acc.} \\ 
\hline 
\multicolumn{4}{|l|}{\textbf{Basic FER Methods}} \\ 
\hline 
Deep learning using CNN and inception blocks \cite{ref:6} & CK+ & 7 & 0.9320 \\ 
\hline 
Manual feature extraction using Local Binary Patterns \cite{ref:8} & CK+ & 7 & 0.9626 \\ 
\hline 
Deep neural network using multi-step pre-processing and feature extraction \cite{ref:9} & CK+ & 7 & 0.9680 \\ 
\hline 
Deep network with joint fine-tuning \cite{ref:10} & CK+ & 7 & 0.9725 \\
\hline 
Boosted deep belief network \cite{ref:11} & CK+ & 7 & 0.9670 \\
\hline
\multicolumn{4}{|l|}{\textbf{Complex FER Methods}} \\
\hline 
Deep CNN using Optical Flow feature extraction \cite{ref:12} & CMED & 18 & 07018 \\ 
\hline 
Deep CNN using knowledge distillation of AUs and SVM classifier \cite{ref:13} & CASME & 5 & 0.8180 \\
\hline 
Continual learning using knowledge distillation \cite{ref:16} & CFEE & 21 & 0.7382 \\
\hline 
Incremental active learning from sparsely annotated data \cite{ref:19} & CFEE & 21 & 0.8502 \\ 
\hline 
SVM classifier using extracted shape and appearance features \cite{ref:17} & CFEE & 21 & 76.91 \\ 
\hline 
\end{tabular}
\caption{State-of-the-art methods for basic and complex FER}
\label{tab:1}
\end{table}

However, human beings express a wide range of emotions through facial expressions that do not fit into predefined categories, and there is evidence that no such basic, prototypical emotion categories exist \cite{ref:20}. Instead, FER develops naturally over time in humans, who are able to identify new, complex emotions on the fly as they appear \cite{ref:1, ref:16}. To approach human-like FER performance, a machine should be able to recognise complex expressions of emotion such as \textit{happily disgusted}, and distinguish them from similar emotions like \textit{happy}, \textit{disgusted}, or \textit{happily surprised}. These \textit{compound expressions} are more than the sum of their parts; they are distinct concepts which express a unique emotion \cite{ref:17}. To humans, such synthesising of known concepts to form new ones comes relatively naturally, and we are able to learn, process and recognise new compound expressions using very little data. However, the state-of-the-art in FER still has difficulty with such cognitive complexity, due to the similarity of features across both basic and compound expressions. AI performs significantly worse at complex FER compared to basic FER, as seen in Table \ref{tab:1}.

\begin{figure}[hbtp]
\centering
\includegraphics[width=\columnwidth]{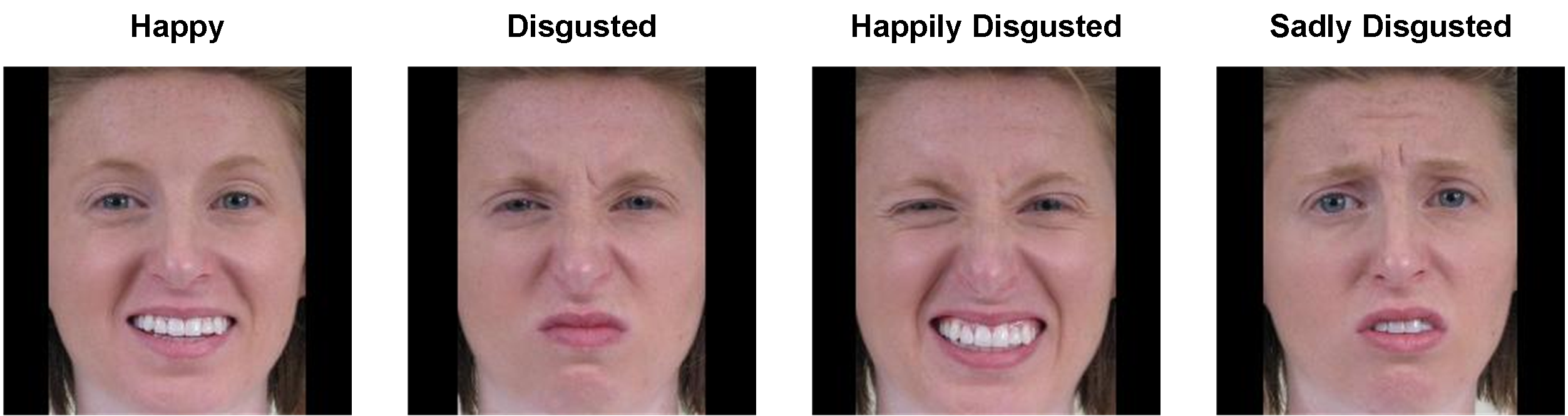}
\caption{Examples from CFEE database \cite{ref:17}. Compound expressions such as \textit{happily disgusted} are more than the sum of their parts (\textit{happy} and \textit{disgusted}).}
\label{fig:1}
\end{figure}

\subsection{FER System Architectures and Design}
\label{sec:2.2}

Complex facial expressions can be represented in a number of ways, such as combinations of basic expressions like \textit{happily surprised} (called \textit{compound expressions}), combinations of expressions in sequence such as \textit{surprise becoming anger}, or as new expressions formed from adding a temporal dimension to basic ones, such as \textit{depression}, which can be categorised as a persistent sadness over time \cite{ref:21}. The \textit{Facial Action Coding System (FACS)} \cite{ref:3} defines 44 \textit{action units (AUs)}, each one representing the activation of particular facial muscles groups. For example, a \textit{happy} expression may be composed of \textit{AUs 12 (lip corner puller)} and \textit{25 (lips part)} since this expression often involves the activation of these muscle groups, whereas a disgusted expression may be composed of \textit{AUs 10 (upper lip raiser)} and \textit{17 (chin raiser)}. The compound expression \textit{happily disgusted} predominantly uses \textit{AUs 10, 12 and 25}, which is the intersection of \textit{AUs} found in each of the basic expressions \textit{happy} and \textit{disgusted} \cite{ref:17}. Our work focuses solely on compound FER as a form of complex FER.

Some traditional FER methods use manual feature extraction techniques such as local binary patterns \cite{ref:8, ref:22}, however deep learning is currently the most popular method, enabling automatic feature extraction through gradient descent backpropagation, and achieving the current state-of-the-art in FER (see Table \ref{tab:1}). High-performing deep learning architectures for computer vision such as \textit{ResNet} \cite{ref:23} and \textit{Xception} \cite{ref:24} can be used by pre-training a model on a large general image database like \textit{ImageNet} \cite{ref:25}, with fine-tuning to adapt to the FER task. This technique is used in a few state-of-the-art models such as \cite{ref:13} but is currently underutilised.

There are a number of common problems in attaining high accuracy for complex FER. Machine learning, particularly deep learning, requires a huge amount of training data to avoid over-fitting, however currently facial expression databases are not sufficient for this task compared with those used for general image recognition or object detection, such as ImageNet \cite{ref:25}. Subject identity bias  (differences in personal attributes of the subjects used for sample images, such as age, gender, ethnic background and level of expressiveness) reduces generalisation performance since there is increased bias towards recognising specific subjects or features (e.g. skin colour or wrinkles) in the training data. Many state-of-the-art FER methods employ various pre-processing methods to reduce the effect of these issues and increase performance.

Face detection and alignment can increase FER performance by creating consistency in training images. For example, the eyes are generally in the same region for images processed by face detection, allowing spatial models such as Convolutional Neural Networks (CNNs) to more easily learn these features. Some common facial detection and alignment techniques include the Viola-Jones face detector \cite{ref:26} as used by \cite{ref:22, ref:27}, generative object detection \cite{ref:28} as used by \cite{ref:29}, supervised descent \cite{ref:30} as used by \cite{ref:6, ref:22}, and the RetinaFace face detector \cite{ref:31}.

Data augmentation applies random transformations to the training image data, such as rotation, flipping and contrast adjustment, generating new images with the same semantic content but different data values. This alleviates the problem created by using smaller FER datasets for deep learning and improves generalisation performance, as the model is trained on more varied data. Data augmentation is used in some FER methods, such as \cite{ref:6, ref:10, ref:19}, but is not used in many state-of-the-art methods, limiting their performance.

\subsection{Few-shot Learning and Continual Learning}
\label{sec:2.3}

As discussed in Section \ref{sec:2.1}, to improve the performance of complex FER systems, a novel approach is required which can incrementally learn new complex expressions as they appear, in the same way that a human might induce the expression \textit{happily surprised} from its prior knowledge of \textit{happy} and \textit{surprised}, using only a few new examples.

Continual learning is a group of methods which focuses on the problem of incrementally adding classes to a trained model. Most of these methods suffer from catastrophic forgetting, which is a reduction in performance on previously learned classes due to substantial weight changes when learning  new ones \cite{ref:16}. Recent advances in continual learning applied to other domains have partially solved the problem of catastrophic forgetting using such methods as memory replay \cite{ref:32} and knowledge distillation \cite{ref:33}. Some pioneering methods of complex FER such as \cite{ref:16, ref:19, ref:34} also use continual learning.

Few-shot learning is a research area focused on training machine learning models with a very small amount of training data, often down to a single sample, making them more adaptable to real-world applications such as streaming video data, security footage where the subject appears only briefly, or learning from a single passport or identification photo supplied by a user. Applying few-shot learning to the domain of complex FER could have a number of applications in areas such as human-computer interaction.

For example, a model that is able to recognise new complex expressions from only a few labelled images could learn a human's emotional state that it hasn't been trained on previously, if they state what they're feeling whilst looking at the machine's camera input. By capturing the facial expression images with some additional processing such as speech recognition, a machine that achieves high performance using few-shot learning could learn this new expression on-the-fly and recognise it the next time it sees it. This approach could enable AI systems to perform better in applications of human-computer interaction such as AI assistants and robotic nurses, which would require human-like levels of emotional intelligence in communication.

The main problem with few-shot learning is under-fitting, whereby the model does not have enough varied training examples to generalise well on recognising unknown samples. A number of techniques are used to increase few-shot learning performance in state-of-the-art methods. \cite{ref:35} achieves high accuracy in one-shot learning for facial identity recognition using supervised auto-encoders to augment a single training sample, producing variations in illumination, expression, occlusion and pose. \cite{ref:36} uses a hybrid feature enhancement network to increase the importance of low-level features such as image textures for semantic segmentation tasks. \cite{ref:37} uses a pseudo-Siamese network to learn classes from a new domain by training one model branch on the original images, and another branch on augmented sketch map images produced by extracting the images' contour features. This method also cites as inspiration the ability of humans to build on prior knowledge when learning new tasks. Common themes amongst these methods of few-shot learning are data augmentation and enhanced feature extraction.

\subsection{Knowledge Distillation}
\label{sec:2.4}

\textit{Knowledge distillation} \cite{ref:33} is a method for transferring knowledge from a teacher model to a student model. The student uses the predicted probabilities from the teacher's output as a soft target in place of the usual hard target (ground truth labels) when calculating the loss between predicted and target outputs. Knowledge is distilled by the student model through gradient descent back-propagation of the distillation loss between the predictions and soft teacher targets. The student thus learns a mapping from its inputs to all the probable label outputs as learned by the teacher, enabling it to quickly form a rich representation of the teacher's knowledge.
 
\begin{figure}[hbtp]
\centering
\includegraphics[width=\columnwidth]{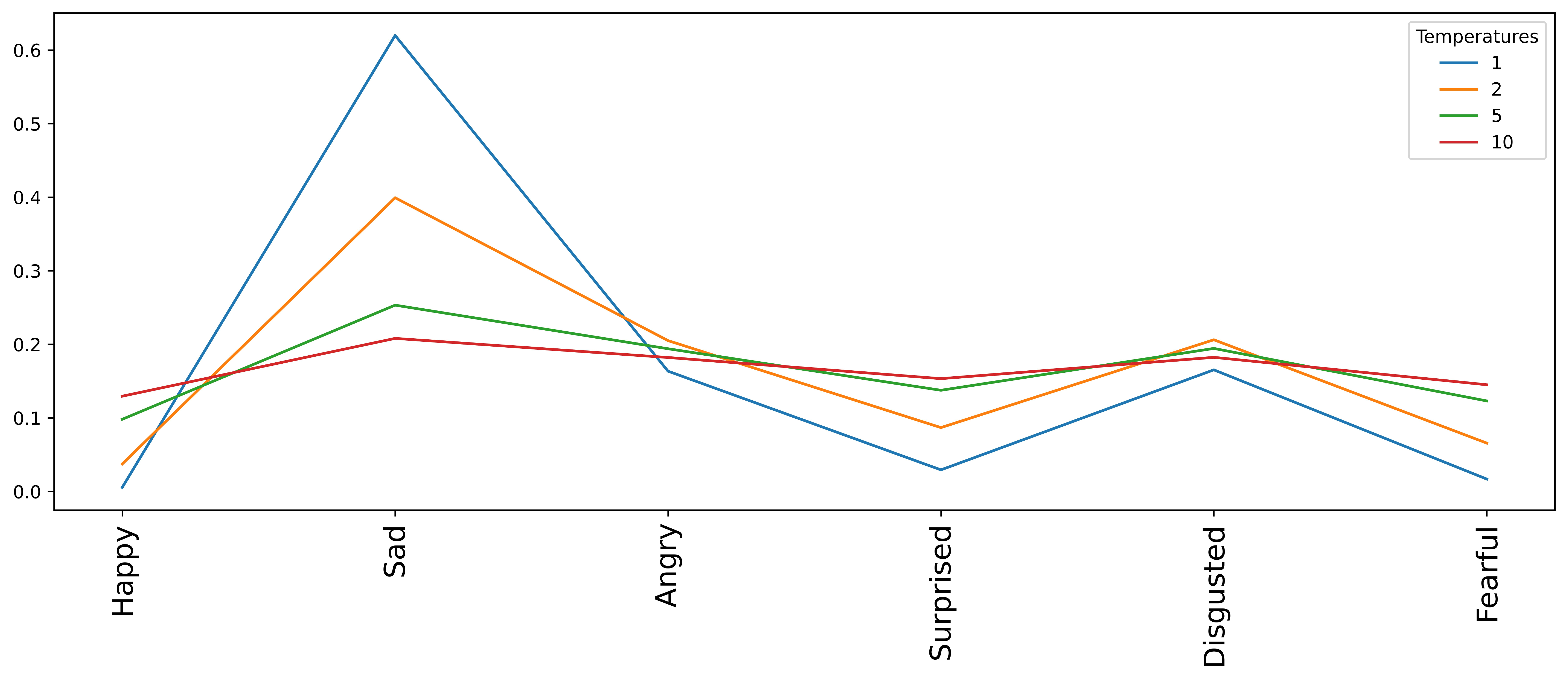}
\caption{Softmax output of 'Surprised' image with different temperatures.}
\label{fig:2}
\end{figure}

The distillation loss is a modified cross-entropy loss whereby the teacher model's prediction output is modified by a temperature $T$ to attain the soft prediction output $y^{soft}$ given by (\ref{eq:1}), where $k$ is the number of classes and $z\in\mathbb{R}^k$ is a vector of logits (the output prior to the softmax activation layer). As seen in Figure \ref{fig:2}, higher temperatures smooth the probability distribution of the soft outputs, giving increased weight to smaller probabilities and decreased weight to the highest probability, whilst retaining the probabilities' relative ranking order.

\begin{equation}
\label{eq:1}
\hat{y}^{soft}=\frac{e^{\frac{z_j}{T}}}{\sum_{j=1}^ke^{\frac{z_j}{T}}}
\end{equation}

(\ref{eq:2}) demonstrates the categorical cross-entropy loss $\mathcal{L}$ used during standard model training, whilst (\ref{eq:3}) is the distillation loss $\mathcal{L}_{dist}$ used for knowledge distillation, whereby $y$ is the true label (one-hot encoded), $\hat{y}$ is the prediction output, and $\hat{y}^{t-soft}$ and $\hat{y}^{soft}$ are the soft teacher prediction and soft student prediction outputs, respectively.

\begin{equation}
\label{eq:2}
\mathcal{L}\left(y,\hat{y}\right)=-\sum_{j=1}^ky_j\log{y_j}
\end{equation}

\begin{equation}
\label{eq:3}
\mathcal{L}_{dist}\left(\hat{y}^{t-soft},\hat{y}^{soft}\right)=-\sum_{j=1}^k\hat{y}^{t-soft}_j\log\hat{y}^{soft}_j
\end{equation}
	
Knowledge distillation can reduce the effects of catastrophic forgetting in continual learning by reinforcing a model's knowledge of known classes when learning new classes. \cite{ref:16} uses knowledge distillation with continual learning for FER through an indicator loss function which is the weighted sum of the distillation loss and hard loss. An indicator function allows the model to treat each new training example differently depending on whether it is reinforcing prior knowledge or gaining new knowledge. For new classes, the hard loss is weighted more and for old classes the distillation loss is weighted more. FER accuracy with this method is reduced significantly less with each new class learned, compared to other methods.

\begin{figure*}[t]
\centering
\includegraphics[width=\textwidth]{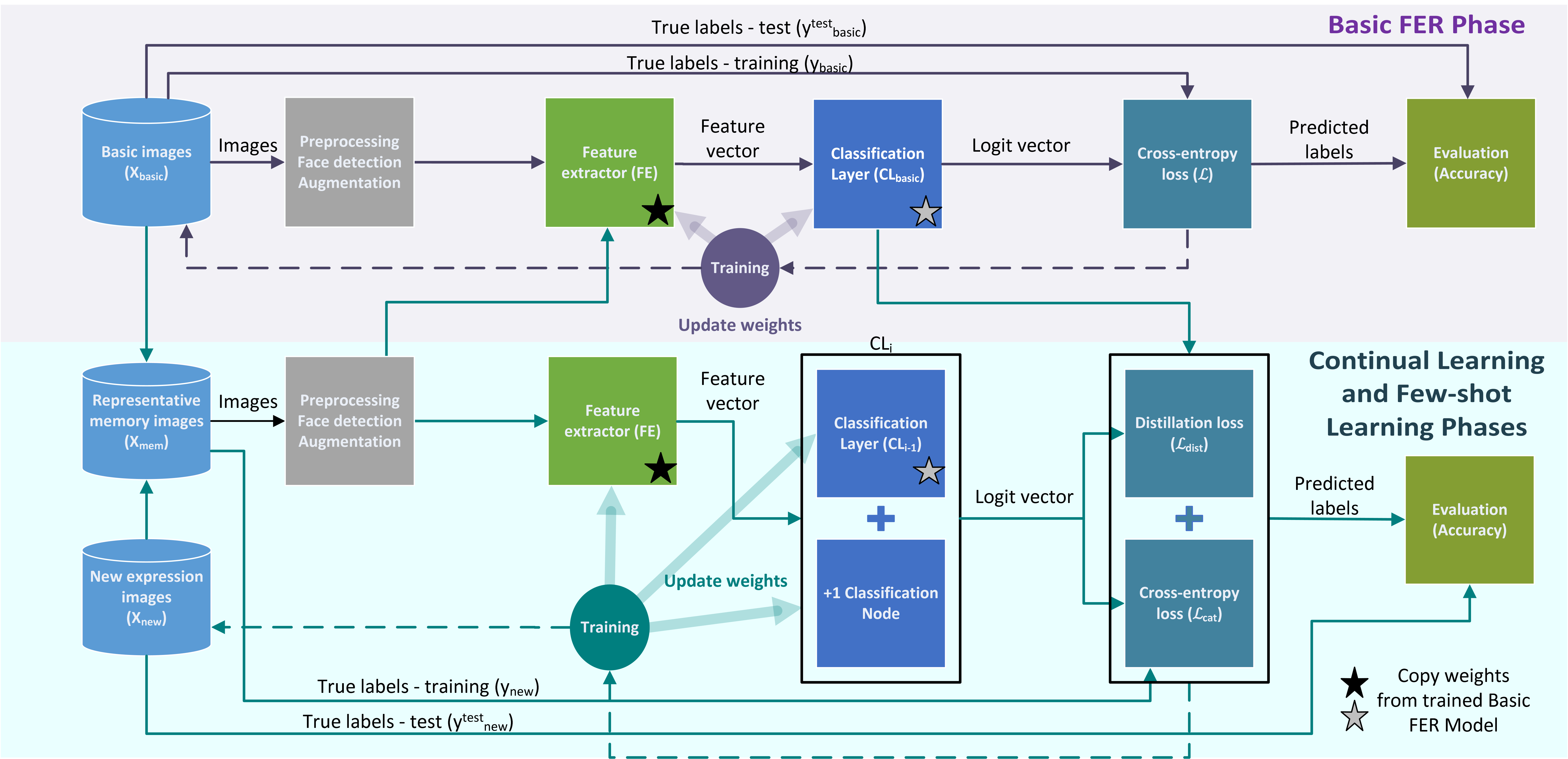}
\caption{System architecture and method}
\label{fig:3}
\end{figure*}

\subsection{Representative Memory Replay}
\label{sec:2.5}

Memory replay techniques are used in continual learning to store a subset of  labelled training examples for known classes, which are added to the new class training data. This helps to prevent catastrophic forgetting by reinforcing the knowledge of known classes when learning new classes. It can also be used in conjunction with knowledge distillation, as in \cite{ref:16, ref:32, ref:38}. Some memory replay methods use specific sample selection techniques with the aim of selecting the most representative samples of each label, such as \cite{ref:32} and \cite{ref:38}. Other methods such as \cite{ref:16} instead use a randomly selected representative memory, stating that randomising the order of inputs when training a model improves generalisation performance. However, a major limitation of this work is that it is not true continual learning, as it draws from the full set of prior training examples using its random selection policy for memory replay. This effectively makes the results comparable to training a new model from the beginning with each new class. Our method aims to improve on \cite{ref:16} whilst adhering to the true principle of continual learning, that is in not having access to the prior training dataset except for a subset of data that is set aside and stored in a representative memory. We use a novel \textit{Predictive Sorting Memory Replay} to select the samples which are most representative of their class. This enhances the knowledge distillation, thereby greatly reducing catastrophic forgetting and improving continual learning performance.

One of the advantages of continual learning is the potential for training a model to learn new classes in real-time without progressively using more resources. To this end, the representative memory is often kept to a fixed number of samples, $K$, over each continual learning iteration, as in \cite{ref:16} and \cite{ref:38}. In this way $m = \frac{K}{k}$ samples are retained for each new class, where $k$ is the number of observed classes so far. However, $k$ increases with each continual learning iteration which progressively reduces the number of samples of each known class in the representative memory, and can reintroduce the effect of catastrophic forgetting. If instead $K$ is allowed to increase, the representative memory can keep a constant number of samples for each label, reducing the effect of catastrophic forgetting at the expense of an ever-growing memory size. \cite{ref:32} evaluates both these methods and compares the results, which as expected shows a decrease in performance using a fixed $K$ compared with an increasing $K$.

\section{Research Design and Methodology}
\label{sec:3}

In this section, the main components of a novel system architecture and method for compound FER are described. The code to reproduce this method is available at \url{https://github.com/AngusMaiden/complex-FER}. The proposed method has three phases:

\begin{itemize}
\item A \textit{Basic FER Phase} in which a \textit{Basic FER Model} learns to recognise six basic expression classes from a dataset of labelled static images of facial expressions.
\item A \textit {Continual Learning Phase} in which the trained model from the \textit{Basic FER Phase} is used to learn new compound expression classes sequentially, by incrementally adding new classes until all expressions have been learned.
\item A \textit{Few-shot Learning Phase} in which the trained model from the Basic FER Phase is used to learn new compound expression classes with only a very small number of training samples from the new class. In this phase, each new class is trained and tested as a separate experiment, re-initialising the model each time based on the model from the Basic FER Phase.\end{itemize}

The re-use of the trained Basic FER model in the \textit{Continual Learning} and \textit{Few-shot Learning Phases} is intended to somewhat mimic the human pattern of learning as we age, by learning complex concepts like compound facial expressions only after the basic concepts, i.e. basic expressions, are understood.

\subsection{Basic FER Phase}
\label{sec:3.1}

\begin{figure*}[t]
\centering
\includegraphics[width=\textwidth]{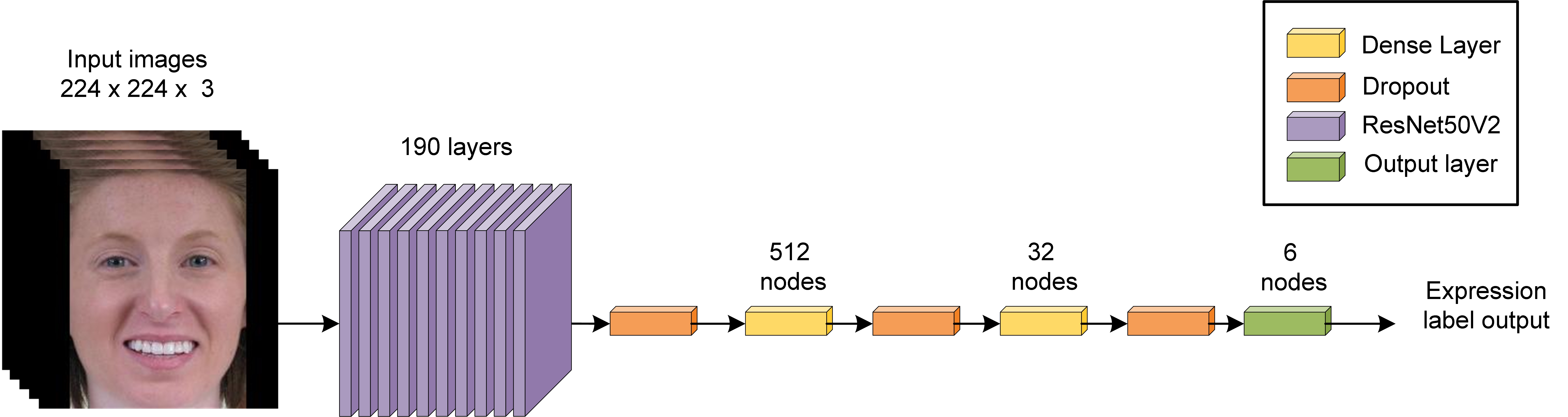}
\caption{Basic FER model architecture}
\label{fig:4}
\end{figure*}

In this initial phase, a Basic FER Model learns the mapping between input images $X_{basic}$ and their ground truth expression labels $y_{basic}\in\mathbb{R}^{k_{basic}}$. Each image is labelled with one of $k_{basic}$ facial expression labels. The model has two main components:

\begin{enumerate}
\item A \textit{feature extractor} $FE$ using a base residual network, \textit{ResNet50V2} \cite{ref:39}, learns feature mappings of images through stacked convolutional blocks with residual connections and dense layers on top.  The network is pre-trained on ImageNet \cite{ref:25} to extract common image features such as shapes and lines, then fine tuned on the FER dataset. The network's output is a feature vector $FE(X_{basic})\in\mathbb{R}^{k_{FE}}$ where $k_{FE}$ is the number of output nodes of $FE$.
\item A classification layer $CL_{basic}$ with $k_{basic}$ output nodes. This layer takes as input the output of $FE(X_{basic})$, and outputs a logit vector $z\in	\mathbb{R}^{k_{basic}}$, where $k_{basic}$ is the number of basic expressions from the Basic FER Phase.
\end{enumerate}

A forward pass of the Basic FER Model from inputs $X_{basic}$ to logit outputs $z$ is given by (\ref{eq:4}). A standard softmax activation function as in (\ref{eq:1}), where $T=1$, is applied to the logit vector to produce the probability vector $\hat{y}\in\mathbb{R}^{k_{basic}}$.

\begin{equation}
\label{eq:4}
z=CL_{basic}(FE(X_{basic}))
\end{equation}

A categorical cross-entropy loss $\mathcal{L}_{cat}$ calculates the error between the predicted labels $\hat{y}$ and ground truth labels $y_{basic}$ for each input image, and is calculated according to (\ref{eq:2}).

The main aim of the \textit{Basic FER Phase} is to train a model to achieve high FER accuracy with robust feature mappings from images of basic facial expressions. The feature mappings are then transferred using knowledge distillation to the models in the \textit{Continual Learning} and \textit{Few-shot Learning Phases}. It is hypothesised that compound FER accuracy can be improved using minimal new training examples through knowledge distillation of basic expressions, as compound and basic facial expressions share some basic features.

In order to achieve a high recognition accuracy in the Basic FER Phase, and subsequently in the \textit{Continual Learning} and \textit{Few-shot Learning Phases}, a good model architecture is needed. Based on empirical knowledge from state-of-the-art research in computer vision, the architecture of the Basic FER Model is designed as displayed in Figure \ref{fig:4}.

The input pipeline for the Basic FER model takes 3-channel (RGB) images of size 224 x 224 x 3 in batches. Each image is labelled with one of $k_{basic}$ basic expression classes. Images are pre-processed using the RetinaFace face detection algorithm \cite{ref:31}. Each image is then normalised such that their pixel values lie between -1 and 1. Data augmentation of the training set is also used to generate random transformations including horizontal flipping, translation and zooming for each image.

\begin{algorithm}
\caption{Basic FER Phase}\label{algo:1}
\begin{algorithmic}[1]
\REQUIRE $X_{basic} \leftarrow$ training images
\REQUIRE $y_{basic} \leftarrow$ labels for images in $X_{basic}$
\REQUIRE $X_{basic}^{test}$ test images
\REQUIRE $y_{basic}^{test}$ labels for images in $X_{basic}$
\REQUIRE $FE \leftarrow ResNet50V2$ (pre-trained on ImageNet)
\REQUIRE $CL_{basic} \leftarrow$ Classification layer
\FOR{epoch in no. of epochs}
\FOR{batch in Batch($X_{basic}$,$y_{basic}$)}
\FOR{x, y in batch}
\STATE $x \leftarrow FaceDetection(x)$
\STATE $x \leftarrow Normalize(x)$
\STATE $x \leftarrow Augment(x)$
\STATE $z \leftarrow CL_{basic}(FE(x))$
\STATE $\hat{y} \leftarrow Softmax(z)$
\STATE $\mathcal{L}_{cat} \leftarrow CrossEntropy(\hat{y},y)$
\ENDFOR
\STATE $\mathcal{L}_{cat} \leftarrow Average(\mathcal{L}_{cat})$ over $batch$
\STATE Update model weights (back-propagation of $\mathcal{L}_{cat}$)
\ENDFOR
\FOR{$x^{test}$, $y^{test}$ in $X_{basic}^{test}$, $y_{basic}^{test}$}
\STATE $\hat{y}^{test} \leftarrow Softmax(CL_{basic}(FE(x^{test}))$
\STATE $acc \leftarrow Accuracy(\hat{y}^{test},y^{test})$
\ENDFOR
\STATE Accuracy = $Average(acc)$
\ENDFOR
\end{algorithmic}
\end{algorithm}

\subsection{Continual Learning Phase}
\label{sec:3.2}

The \textit{Continual Learning Phase} is an iterative cycle whereby labelled images from each complex expression class are added to the existing dataset and trained sequentially. At each iteration $i$, one new complex expression class is selected, which is comprised of images $X_{new_i}$ and associated true labels $y_{new_i}\in\mathbb{R}^{k_i}$, where $k_i$ is the total number of expression classes available at this iteration ($k_i$=$k_{basic}$+$i$). The total number of known (trained) classes at the beginning of each iteration is $k_{i-1}$. The phase runs for $k_{compound}$ iterations, whereby $k_{compound}$ is the number of compound expression classes available. For each iteration $i$, a new node is added to the classification layer of the previous model, $CL_{i-1}$, with the new layer denoted as $CL_i$. This new node's weights are randomly initialised using a Glorot Uniform distribution, whilst the other $k_{i-1}$ nodes inherit their weights from the corresponding $CL_{i-1}$ layer of the trained model from the previous iteration $i-1$. In the first iteration, whereby $i=1$, the previous model is the trained model from the Basic FER Phase, such that $CL_0$=$CL_{basic}$. The structure of the feature extractor $FE$ does not change at each iteration and is reused as-is. Figure \ref{fig:3} provides a visualisation of the continual learning method and process flow.

A representative memory $X_{mem_i}$ is used to store a portion of training samples from the previous iteration $i-1$ together with the new class training samples $X_{new_i}$. A number of training samples $X_{select_i}$, and their associated labels $y_{select_i}$, are selected according to the Predictive Sorting Memory Replay (PSMR) selection policy. The pseudo-code for this policy is outlined in Algorithm \ref{algo:2}.

\begin{algorithm}
\caption{Predictive Sorting Memory Replay (PSMR)}\label{algo:2}
\begin{algorithmic}[1]
\REQUIRE $K \leftarrow$ number of samples in representative memory
\REQUIRE $k_i \leftarrow$ number of classes at iteration i
\REQUIRE $m \leftarrow \frac{K}{k_{i-1}}$
\REQUIRE $X_{mem_{i-1}}, y_{mem_{i-1}} \leftarrow$ representative memory of previous iteration
\IF{$i=1$}
\STATE $X_{select_i},y_{select_i} \leftarrow$ randomly select $K$ $X_{basic}$ images and associated labels
\ELSE
\STATE $\hat{y} \leftarrow Softmax(CL_i(FE(X_{mem_{i-1}}))$
\FOR{each $\hat{y}_j$ in $j$ classes where $j=1,...,k_i$}
\STATE $Sort(\hat{y}_j)$ in order of prediction probability
\STATE $Append(y_{select_i}) \leftarrow$ Select top $m$ sorted labels $\hat{y}_j$
\STATE $Append(X_{select_i}) \leftarrow$ images for $y_{select_i}$ labels
\ENDFOR
\ENDIF
\RETURN $X_{select_i}, y_{select_i}$
\end{algorithmic}
\end{algorithm}

Using this policy, the representative memory is therefore comprised of images that are the most representative of their respective class. When learning new classes, these representative memory samples are trained alongside the new class samples, which reinforces the knowledge of previous classes and reduces the effect of catastrophic forgetting. In a similar way, humans retain memories of only the most pronounced moments of an experience, which efficiently enables recognition and classification of the entire experience.
Once initialised, the representative memory does not acquire any new samples from the \textit{Basic FER Phase} data, to emulate human learning whereby the raw data from prior experiences is no longer available, and only memories are retained. This also ensures the method is aligned with practical applications that may have limiting memory and computation requirements, such as mobile computing, IoT and robotics.

A forward pass of the continual learning model at iteration $i$ from inputs $X_{mem_i}$ to output logits $z_i$ is given in (\ref{eq:5}). A softmax activation function as in (\ref{eq:1}), where $T=1$, is applied to the logit vector to produce the prediction vector $\hat{y}_i\in\mathbb{R}^{k_i}$.

\begin{equation}
\label{eq:5}
z_i=CL_i(FE(X_{mem_i}))
\end{equation}

\subsection{Distillation Loss}
\label{sec:3.3}

The loss function of a neural network calculates the difference between the target output, usually the ground truth labels, and the predicted output of a model. By minimising the loss through gradient descent optimisation, we aim to minimise this difference, bringing the predicted outputs closer to the real outputs with each gradient descent step. In this phase, the loss $\mathcal{L}$ is the weighted sum of a standard categorical cross-entropy loss $\mathcal{L}_{cat}$ and distillation loss $\mathcal{L}_{dist}$, with a distillation weighting factor, $\gamma:0\leq\gamma\leq1$, as shown in (\ref{eq:6}).

\begin{equation}
\label{eq:6}
\mathcal{L}(y_i,\hat{y}_i)=(\gamma-1)\cdot\mathcal{L}_{cat}(y_i,\hat{y}_i)+\gamma\cdot\mathcal{L}_{dist}(\hat{y}_i^{t-soft},\hat{y}_i^{soft})
\end{equation}

The cross-entropy loss $\mathcal{L}_{cat}$ calculates the error between the predicted labels $\hat{y}_i$ and ground truth labels $y_i$ for each input image, as in (\ref{eq:2}). The distillation loss $\mathcal{L}_{dist}$ calculates the error between $\hat{y}_i^{t-soft}$and $\hat{y}_i^{soft}$, as in (\ref{eq:3}). Our method demonstrates a unique handling of the distillation loss whereby the teacher model is a static copy of the trained model attained in the \textit{Basic FER Phase}. The weights of this teacher model are never updated in the Continual Learning Phase. A forward pass of the teacher model at iteration $i$ from inputs $X_{mem_i}$ to teacher logits $z_i^t$ is given in (\ref{eq:7}). A softmax activation function with temperature $T$, as in (\ref{eq:1}), is applied to the logits $z_i$ from (\ref{eq:5}) to produce student predictions $\hat{y}_i^{soft}$ and to the teacher logits $z_i^t$ from (\ref{eq:7}) to produce teacher predictions $\hat{y}_i^{t-soft}$, which are used in the calculation of the distillation loss $\mathcal{L}_{dist}$ (\ref{eq:6}).

\begin{equation}
\label{eq:7}
z_i^t=CL_{basic}(FE_{basic}(X_{mem_i}))
\end{equation}

With each iteration, $\mathcal{L}_{dist}$ naturally increases, adding a higher penalty to the overall loss $\mathcal{L}$. This is because $\mathcal{L}_{dist}$ is calculated between the vectors $\hat{y}_{t-soft}\in\mathbb{R}^{k_{basic}}$ (zero-padded) and $\hat{y}_i^{soft}\in\mathbb{R}^{k_i}$. Due to the zero-padding, there will be $i$ constant zero values in $\hat{y}$, causing $\mathcal{L}_{dist}$ to increase with each step $i$. To counteract this effect, a distillation weight decay is used, as in (\ref{eq:8}), which decreases the value of $\gamma$ with each iteration $i$, thereby reducing the  weight of $\mathcal{L}_{dist}$ in the overall loss $\mathcal{L}$ (\ref{eq:6}). This function approaches but never equals the asymptote at $\gamma=0$, ensuring there is always some amount of distilled knowledge of basic expressions contributing to the overall loss.

\begin{equation}
\label{eq:8}
\gamma_i=\gamma_{i-1}e^{\frac{-1}{1+e}}
\end{equation}

After each epoch of training, the accuracy of the continual learning model is tested using a held-out test dataset. Training is stopped when the test accuracy is no longer improving over previous epochs. We then record the test accuracy for iteration $i$ for all $k_i$ trained expression classes, as well as the single class test accuracy for the newest facial expression. Evaluating this method of knowledge distillation in the \textit{Continual Learning} and \textit{Few-shot Learning Phases} tests the hypothesis that compound FER accuracy can be improved using few new training examples through knowledge distillation of basic expressions.

\begin{algorithm}
\caption{Continual Learning Phase}\label{algo:3}
\begin{algorithmic}[1]
\REQUIRE $X_{basic} \leftarrow$ basic training images
\REQUIRE $y_{basic} \leftarrow$ labels for images in $X_{basic}$
\REQUIRE $X_{new} \leftarrow$ compound training images
\REQUIRE $y_{new} \leftarrow$ labels for images in $X_{new}$
\REQUIRE $X_{new}^{test} \leftarrow$ compound test images
\REQUIRE $y_{new}^{test} \leftarrow$ labels for images in $X_{new}$
\REQUIRE $FE \leftarrow ResNet50V2$ (trained in Basic FER Phase)
\REQUIRE $CL_0 \leftarrow CL_{basic}$ (from Basic FER Phase)
\REQUIRE $T \leftarrow$ Softmax Temperature for Distillation Loss
\REQUIRE $\gamma \leftarrow$ distillation weighting factor
\FOR{i in list of $k_{compound}$ facial expressions}
\STATE $X_{new_i}, y_{new_i} \leftarrow$ select images, labels of next expression in list
\STATE $y_{select_i}, X_{select_i} \leftarrow PSMR$
\STATE $X_{mem_i} \leftarrow Concatenate(X_{select_i},X_{new_i})$
\STATE $y_{mem_i} \leftarrow Concatenate(y_{select_i},y_{new_i})$
\STATE $CL_i \leftarrow$ Add one output node to $CL_{i-1}$
\FOR{epoch in no. of epochs}
\FOR{$batch$ in Batch($X_{mem_i}$, $y_{mem_i}$)}
\FOR{$x$, $y$ in $batch$}
\STATE $x \leftarrow FaceDetection(x)$
\STATE $x \leftarrow Normalize(x)$
\STATE $x \leftarrow Augment(x)$
\STATE $z \leftarrow CL_i(FE(x))$
\STATE $\hat{y} \leftarrow Softmax(z)$
\STATE $z^t \leftarrow CL_{basic}(FE_{basic}(x))$
\STATE $\hat{y}^{t-soft} \leftarrow Softmax(z^t, T)$
\STATE $\hat{y}^{soft} \leftarrow Softmax(z, T)$
\STATE $\mathcal{L}_{cat} \leftarrow CrossEntropy(\hat{y},y)$
\STATE $\mathcal{L}_{dist} \leftarrow CrossEntropy(\hat{y}^{t-soft},\hat{y}^{soft})$
\STATE $\mathcal{L} \leftarrow (\gamma-1)\cdot\mathcal{L}_{cat}+\gamma\cdot\mathcal{L}_{dist}$
\ENDFOR
\STATE $\mathcal{L} \leftarrow Average(\mathcal{L})$ over $batch$
\STATE Update model weights (back-propagation of $\mathcal{L}$)
\ENDFOR
\FOR{$x^{test}$, $y^{test}$ in $X_{basic}^{test}$, $y_{basic}^{test}$}
\STATE $\hat{y}^{test} \leftarrow Softmax(CL_i(FE(x^{test}))$
\STATE $acc \leftarrow Accuracy(\hat{y}^{test},y^{test})$
\ENDFOR
\STATE Accuracy = $Average(acc)$
\ENDFOR
\STATE Update weight decay $\gamma$ according to (\ref{eq:8})
\ENDFOR
\end{algorithmic}
\end{algorithm}

\subsection{Few-shot Learning Phase}
\label{sec:3.4}

In this phase, the hypothesis that better complex FER performance can be achieved through knowledge distillation of basic expressions is tested in the context of few-shot learning, whereby a very limited number of training examples are used. We run one experiment for each of the $k_{compound}$ compound facial expressions, with repeat trials of 5, 3 and 1 training examples. To train these expressions, the same system architecture and methodology is used as in the \textit{Continual Learning Phase}, with the exception of the \textit{Representative Memory Replay} component, which is excluded from this phase as it is specific to continual learning. Each few-shot learning experiment is equivalent to one iteration of the continual learning experiment. The model's parameters are reset after each experiment, before a new class is chosen. These experiments test the model's ability to learn each separate complex expression class using very few training examples through distilling the knowledge of basic facial expressions. The aggregate results of these separate experiments allows us to generalise about the model's capabilities for few-shot learning with compound FER.

For this phase, at the beginning of each experiment $j\in[1,\ldots,k_{compound}]$, one new compound expression class is selected which is comprised of images $X_{new_j}$ and associated true labels $y_{new_j}\in\mathbb{R}^{k_{fewshot}}$, where $k_{fewshot}=k_{basic}+1$. As with the first iteration of the \textit{Continual Learning Phase}, the classification layer $CL_{fewshot}$ is comprised of a new node added to the classification layer of the Basic FER model $CL_{basic}$. For each experiment $j$, a forward pass of the few-shot learning model produces the output logits $z_j$ as in (\ref{eq:9}), whilst a forward pass of the teacher model produces the output logits $z_j^t$ as in (\ref{eq:10}). A softmax activation function is applied to the logit vectors $z_j$ and $z_j^t$ to produce the prediction vectors $\hat{y}_j^{soft}$ and $\hat{y}_j^{t-soft}$, respectively, as in (\ref{eq:1}). The cross-entropy loss $\mathcal{L}_{cat}$ calculates the error between the predicted labels $\hat{y}_j$ and ground truth labels $y_j$ for each input image, as in (\ref{eq:2}). The distillation loss $\mathcal{L}_{dist}$ calculates the error between $\hat{y}_j^{t-soft}$ and $y_j^{soft}$, as in (\ref{eq:3}). The loss function is given in (\ref{eq:6}) and calculated as in the \textit{Continual Learning Phase}.

\begin{equation}
\label{eq:9}
z_j=CL_{fewshot}(FE(X_{new_j}))
\end{equation}

\begin{equation}
\label{eq:10}
z_j^t=CL_{basic}(FE_{basic}(X_{new_j}))
\end{equation}

After each epoch of training, the accuracy of the few-shot learning model is tested using the same held-out test dataset from the Continual Learning Phase. Training is stopped when the test accuracy is no longer improving over previous epochs. Few-shot learning experiments are conducted using 5, 3 and 1 training examples per class, with the single class test accuracy and number of training steps recorded for each $j\in\mathbb{R}^{k_{compound}}$ compound expression classes.

\section{Grad-CAM Visualisation of Basic and Compound Features}
\label{sec:3.5}

This experimental setup was also used to visualise the features of basic expressions which are distilled into the model when learning new complex expressions. The \textit{Grad-CAM} \cite{ref:40} method was applied to visualise the areas of each image that are most activated for particular expression classes. Using this method, we can visualise the features of basic expressions that are common to those compound expressions derived from them. Initially, Grad-CAM heat maps were produced from the Basic FER model for the $k_{basic}$ basic expressions. Similarly to the Few-Shot Learning Phase, here we use just one iteration of the Continual Learning Phase, without the representative memory, training the model for $j\in\mathbb{R}^{k_{compound}}$ compound expressions and resetting the weights each time, but using all available training examples in order to produce the strongest feature activations. The trained models were then used to produce Grad-CAM feature maps for each expression which highlights the areas of the image which are most activated when predicting that expression. Connections between the basic and compound expressions were drawn by identifying semantic relationships between them and their Grad-CAM feature maps in order to highlight facial expression features which are common to these expressions.

For example, Figure 5 demonstrates the Grad-CAM visualisation of the basic features \textit{angry}, \textit{disgusted} and \textit{fearful} in relation to the compound expressions \textit{angrily disgusted} and \textit{fearfully angry}. In this visualisation, we can clearly see the features which are common to both the basic and compound expressions, such as in the furrowed brow which is common to the \textit{fearful}, \textit{angry} and \textit{fearfully angry} expressions. We can also see how the \textit{angrily disgusted} expression shares the down-turned mouth with the \textit{disgusted expression} and the furrowed brow with the \textit{angry expression}. These common features also represent the knowledge which is transferred from the Basic FER model to the continual learning model using knowledge distillation. This visualisation, together with the improved performance in the Continual Learning and Few-Shot Learning Phases of the models which use knowledge distillation of basic features, supports the hypothesis that knowledge of basic features can improve the recognition of compound facial expression labels which share those features.

\begin{figure}[!h]
\centering
\includegraphics[width=\columnwidth]{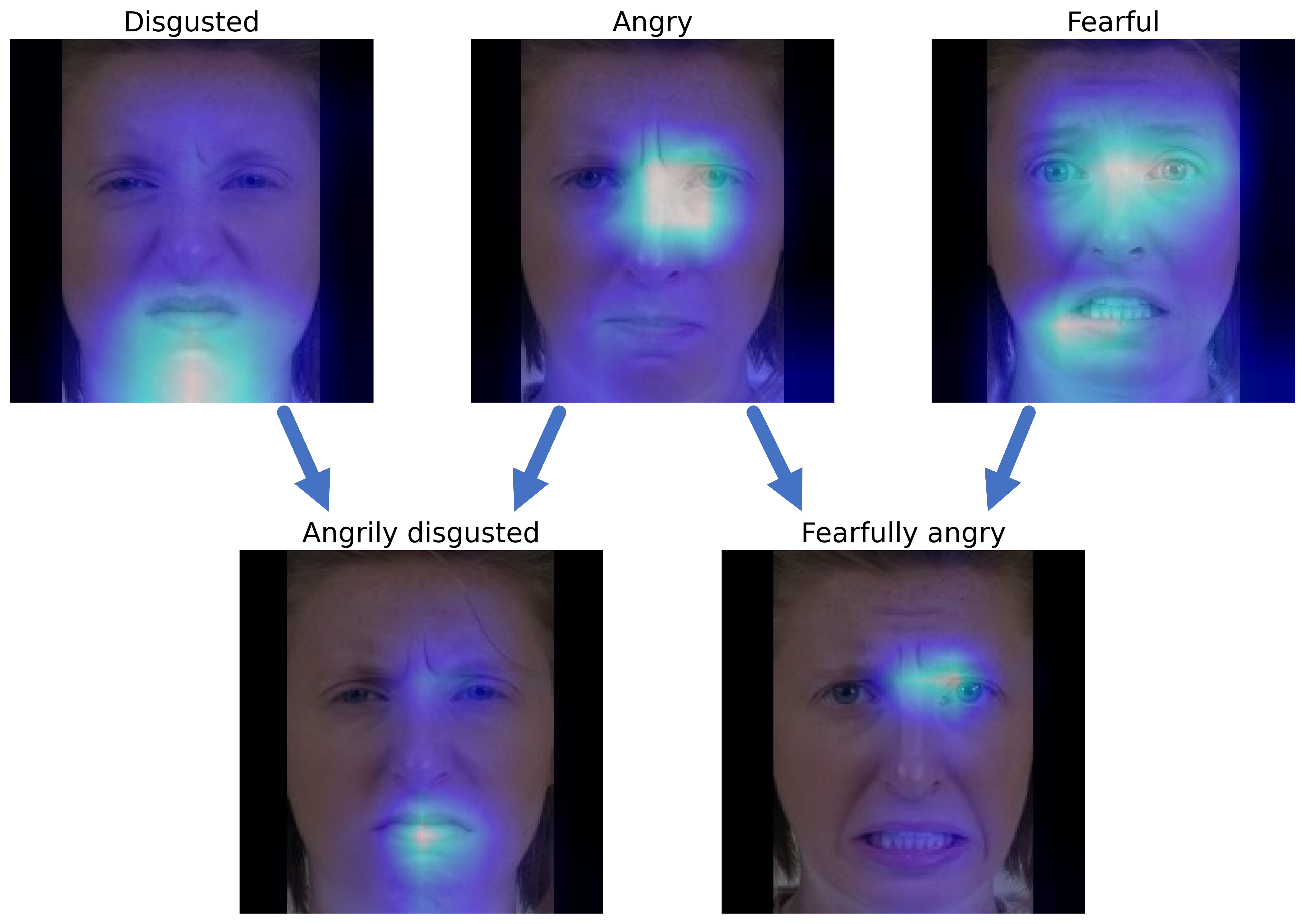}
\caption{Grad-CAM Visualisation of basic features in angrily disgusted and fearfully angry compound expressions}
\label{fig:5}
\end{figure}

\begin{figure}[!h]
\centering
\includegraphics[width=\columnwidth]{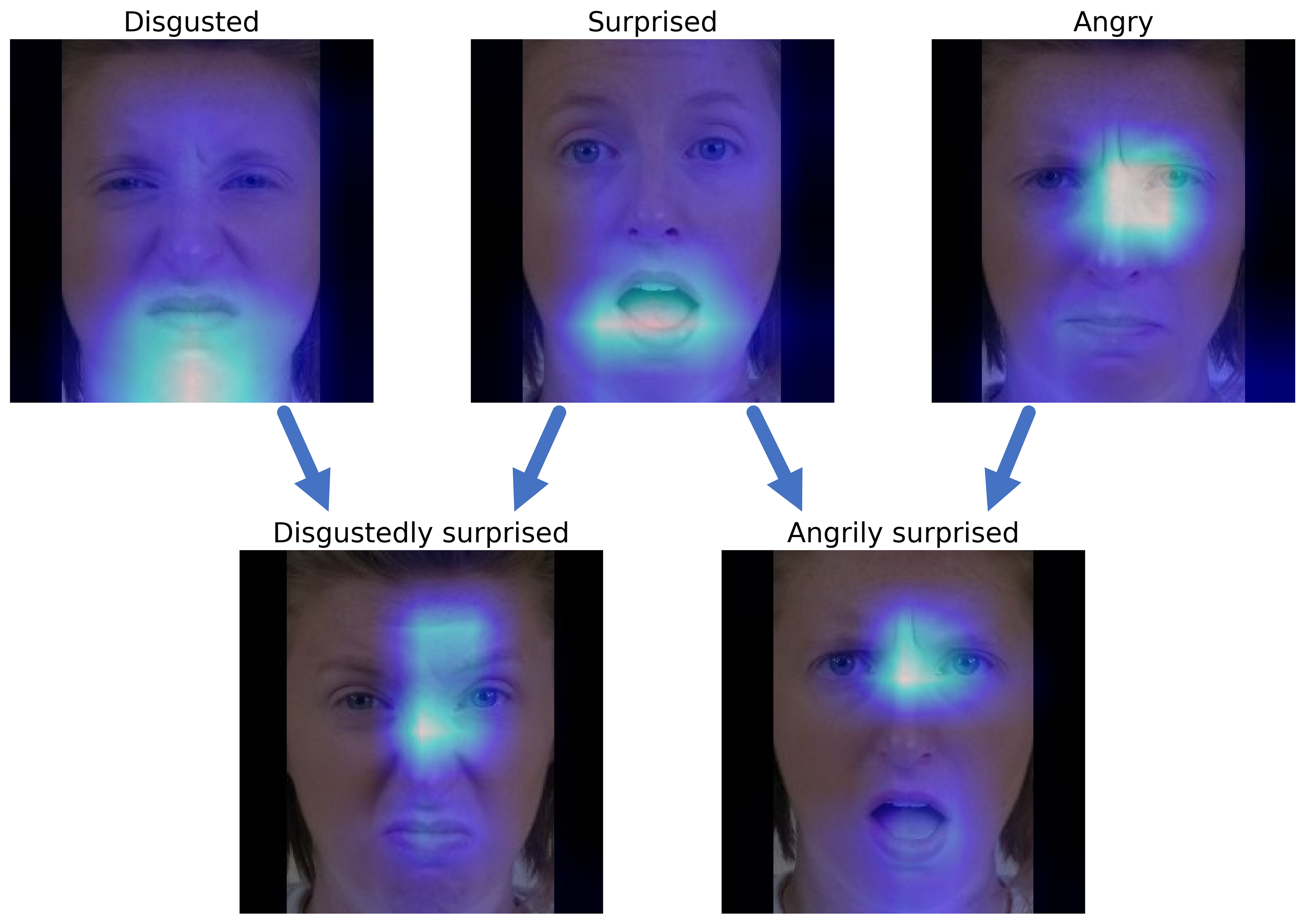}
\caption{Grad-CAM visualisation of basic features in angrily surprised disgustedly surprised and compound expressions}
\label{fig:6}
\end{figure}

\begin{figure}[!h]
\centering
\includegraphics[width=\columnwidth]{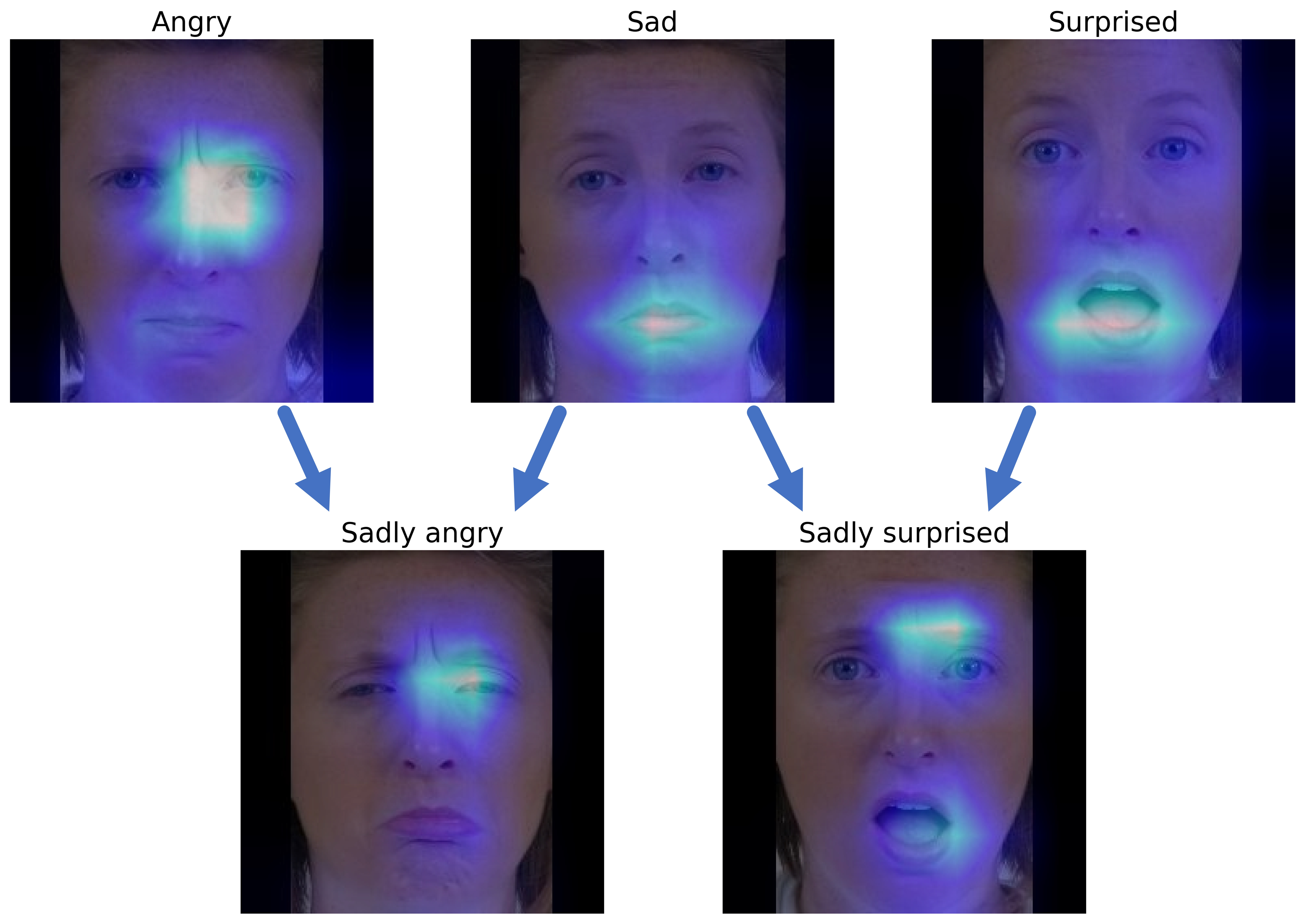}
\caption{Grad-CAM visualisation of basic features in disgustedly surprised and angrily surprised compound expressions}
\label{fig:7}
\end{figure}

\section{Empirical Evaluation and Results}
\label{sec:4}

The dataset used for evaluation is the \textit{Compound Facial Expressions of Emotion (CFEE)} database \cite{ref:17}. It contains  5044 images of 230 different subjects with acted facial expressions in a controlled lab environment. The dataset is labelled with 21 different facial expressions.

For evaluation, a \textit{subject-independent k-fold cross validation} method was used whereby the dataset is partitioned into 10 folds of 23 subjects each. 1 fold is held out as the test set, with the training data comprising the remaining 9 folds. The fold split points are chosen according to subject, such that the test data contains images of completely different subjects than the training data. This aims to reduce the effect of subject identity bias and improve the generalisation performance of the trained model, whilst also simulating the model's performance with real subjects that it has not seen before.

In the \textit{Basic FER Phase}, the model is trained to completion, with the accuracy evaluated using the test dataset at the end of each epoch. The next test fold is then selected, with the training data comprising the remaining 9 folds, and the model is again trained and evaluated. This process is iterated over 10 times for the 10 possible test and training set combinations. The final evaluation results are aggregated to produce a maximum, mean and standard deviation for Basic FER accuracy.

A set of six basic expression classes were chosen in line with \cite{ref:3} (\textit{happy}, \textit{sad}, \textit{angry}, \textit{surprised}, \textit{disgusted}, \textit{fearful}), making $k_{basic}=6$. These basic expression images comprise the $X_{basic}$ images used in the \textit{Basic FER Phase}, and their corresponding labels are the true labels $y_{basic}$. The compound expression classes are the remaining set of 15 classes in the CFEE \cite{ref:17} dataset (\textit{happily surprised}, \textit{happily disgusted}, \textit{sadly angry}, \textit{angrily disgusted}, \textit{appalled}, \textit{hatred}, \textit{angrily surprised}, \textit{sadly surprised}, \textit{disgustedly surprised}, \textit{fearfully surprised}, \textit{awed}, \textit{sadly fearful}, \textit{fearfully disgusted}, \textit{fearfully angry}, and \textit{sadly disgusted}), making $k_{compound}=15$. The total number of classes available in the dataset is $k_{total}=k_{basic}+k_{compound}$.

A ResNet50V2 model \cite{ref:40} is used as the base model for the $FE$ network and pre-trained on \textit{ImageNet} \cite{ref:25} with fine-tuning. Initially, the base model is frozen to train only the top dense layers on the FER task. Early-stopping is used by monitoring the test accuracy during training. Once test accuracy has ceased increasing, training stops and the best model weights are restored. Next, the layers of the base ResNet model are unfrozen and the model is trained again to fine-tune these weights to the FER task. During training in the \textit{Continual Learning} and \textit{Few-shot Learning Phases}, the layers of the first two convolutional blocks in the base model are frozen, since knowledge encoded in these weights relates to the fundamentals of image recognition such as lines and shapes. Freezing these layers prevents the weights from being destroyed when learning new expression classes, and also saves computational power and training time, making the model more practically applicable.

A initial range of good hyperparameters for the Basic FER Model were chosen based on empirical knowledge of good deep learning architectures. These hyperparameters form a multi-dimensional search space which can be searched to optimise the model. The \textit{Hyperband} optimisation algorithm \cite{ref:41} was chosen as the search algorithm and implemented using \textit{KerasTuner} \cite{ref:42}. Hyperband uses an infinite-armed bandit method combined with successive halving to explore the search space and converge to an optimal set of hyperparameters. Following this tuning process, the optimal hyperparameters were selected as in Table \ref{tab:2}.

\begin{table}[!h]
\centering
\begin{tabular}{|l|l|}
\hline 
\multicolumn{2}{|l|}{\textbf{Basic FER Phase}} \\
\hline 
Initial Training epochs	& 1000 (early stopping) \\
\hline
Fine Tuning Training epochs	& 1000 (early stopping) \\
\hline
Batch size & 32 \\
\hline
Optimisation function & Adam \\
\hline
Initial Learning Rate & 1e-4 \\
\hline
Fine Tuning Learning Rate & 1e-6 \\
\hline
\multicolumn{2}{|l|}{\textbf{Continual Learning and Few-shot Learning Phases}} \\
\hline
Training epochs & 1000 (early stopping) \\
\hline
Batch size & 32 \\ 
\hline
Optimisation function & Adam \\
\hline
Learning Rate & 1e-5 \\
\hline
Temperature (T) & 3 \\
\hline
Distillation Weight ($\gamma$) & 0.1 \\
\hline
\end{tabular}
\caption{Hyperparameters}
\label{tab:2}
\end{table}

Evaluating the model in the Basic FER Phase using Subject-Independent K-Fold Cross Validation, a maximum, mean and standard deviation FER accuracy are obtained as shown in Table \ref{tab:3}. The best results are comparable with state-of-the-art basic FER methods (see Table \ref{tab:1}).

\begin{table}[!h]
\centering
\begin{tabular}{|l|l|l|}
\hline
\textbf{Max Acc.} & \textbf{Mean Acc.} & \textbf{SD Acc.} \\
\hline
0.9624 & 0.8525 & 0.061 \\
\hline
\end{tabular}
\caption{Accuracy of Basic FER Model using Subject-Independent K-fold Cross Validation}
\label{tab:3}
\end{table}

For the \textit{Continual Learning Phase}, the best performing test set from the \textit{Basic FER Phase} cross validation was used. The continual learning model was evaluated using the method and metrics developed by \cite{ref:16}. Using this method, the sequence of new classes learned in continual learning are randomised in order to test the model's invariance to the sequence order of expression classes. For each randomised list of complex expressions, out of $C$ total lists (where $C=10$ for our experiments), the accuracy at each continual learning step $i$, over $N$ number of test samples, is recorded. The average step accuracy, $aveSA_i$ is the average accuracy at step $i$ over all lists $C$, as shown in (\ref{eq:11}).

\begin{equation}
\label{eq:11}
aveSA_i=\frac{1}{C}\sum_{j=1}^C\frac{1}{|y^j|}\sum_{n=1}^NH(y_n^j,\hat{y}_n^j)
\end{equation}

Where $H(y_n^j,\hat{y}_n^j)$ is an indicator function that returns $1$ if $y_n^j=\hat{y}_n^j$ and $0$ if $y_n^j\neq\hat{y}_n^j$. Furthermore, the $Overall Accuracy$ is calculated as the average $aveSA$ over all continual learning steps $i$, as demonstrated in (\ref{eq:12}).

\begin{equation}
\label{eq:12}
Overall Accuracy=\frac{1}{k_{total}}\sum_{i=1}^{k_{total}}aveSA_i
\end{equation}

\begin{figure*}[!t]
\centering
\includegraphics[width=\textwidth]{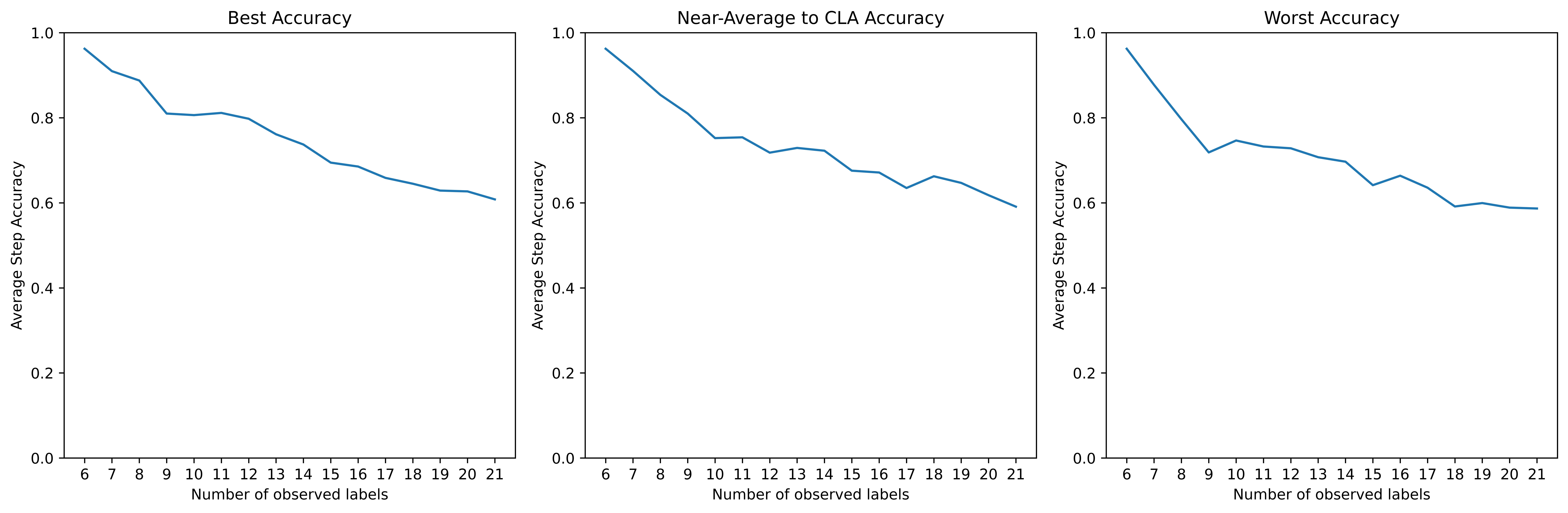}
\caption{Best, worst and near-average accuracy at each continual learning step $i$}
\label{fig:8}
\end{figure*}

\begin{figure*}[!t]
\centering
\includegraphics[width=\textwidth]{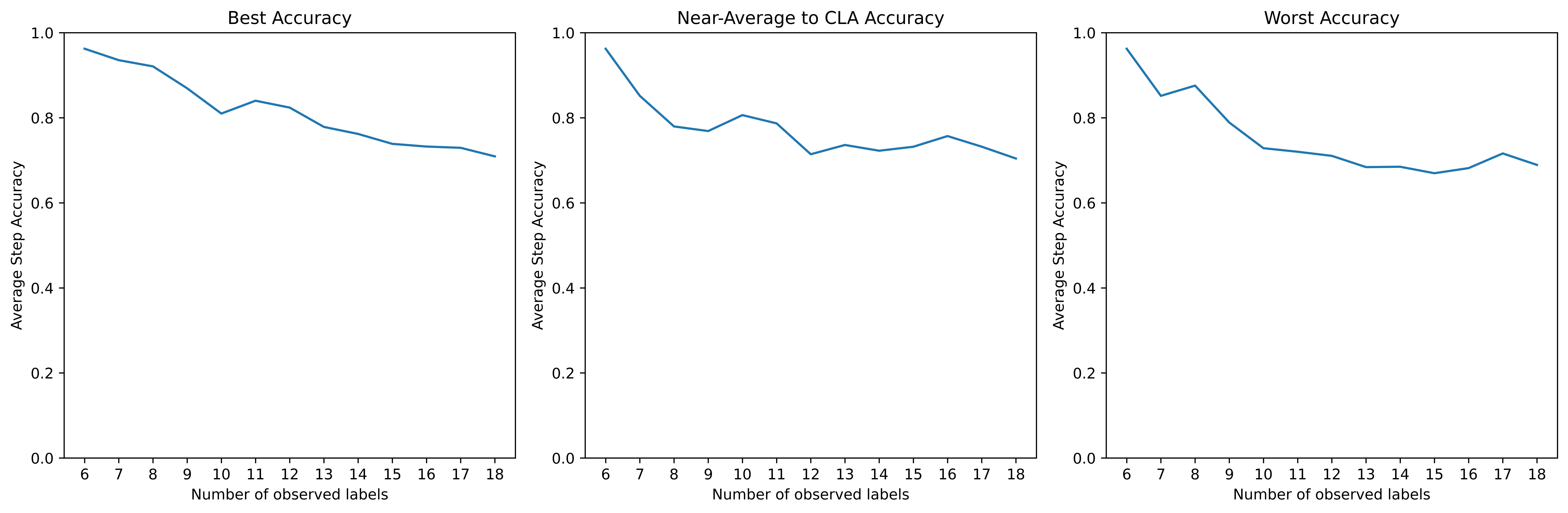}
\caption{Best, worst and near-average accuracy at each continual learning step $i$ (excl. singular labels*)}
\label{fig:9}
\end{figure*}

The $Overall Accuracy$ results obtained in the \textit{Continual Learning Phase} are displayed in Table \ref{tab:4}, along with comparable baseline results as reported in \cite{ref:16}. Furthermore, out of the 10 randomised complex expression lists, Figures \ref{fig:8} and \ref{fig:9} demonstrate the performance of the best, worst, and nearest to $Overall Accuracy$ in terms of the average accuracy over each step $i$.

\begin{table}[!h]
\centering
\begin{tabular}{|l|l|}
\hline
\textbf{Method} & \textbf{Overall Accuracy} \\
\hline
Tree-CNN \cite{ref:45} & (0.5107, 0.6437) \\
\hline
Fine-tuning \cite{ref:54} & (0.6837, 0.7418) \\
\hline
LwF \cite{ref:46}  & (0.5373, 0.6638) \\
\hline
TOPIC \cite{ref:50} & (0.5268, 0.7168) \\
\hline
Deep SLDA \cite{ref:52} & (0.5387, 0.7478) \\
\hline
REMIND \cite{ref:51} & (0.5398, 0.7463) \\
\hline
Lucir-CNN \cite{ref:48} & (0.5698, 0.7639) \\
\hline
PODNet-CNN \cite{ref:49} & (0.5991, 0.8163) \\
\hline
Lucir \cite{ref:48} /w AANets \cite{ref:53} & (0.6414, 0.8598) \\
\hline
PODNet-CNN \cite{ref:49} /w AANets \cite{ref:53} & (0.6781, 0.8697)\\
\hline
iCaRL \cite{ref:47} & (0.7138, 0.8327) \\
\hline
DCLEER \cite{ref:16} & (0.7361, 0.8904) \\
\hline
\textbf{Our Method} & \textbf{(0.7428, 0.7327)} \\
\hline
\textbf{Our Method (excl. singular labels*)} & \textbf{(0.8232, 0.7810)} \\
\hline
\end{tabular}
\caption{Overall Accuracy of our method compared with other continual learning methods \textit{(unknown class only, all classes)}}
\label{tab:4}
\end{table}

*Additionally, a second experiment was conducted using only compound facial expressions whose labels are composed of two basic expression labels, such that the expressions \textit{hatred}, \textit{appalled}, and \textit{awed} are omitted and $k_{compound}=12$. The semantic relationship between these singular expressions and the six basic facial expressions is not as clear as with expressions like \textit{happily surprised}, which is clearly related to  \textit{happy} and \textit{surprised}. By removing these singular expressions, we can focus on FER for compound expressions and more thoroughly test the hypothesis that higher accuracy can be achieved for compound FER through knowledge distillation of the features of basic expressions which are common to related compound expressions.

To further evaluate the use of continual learning for compound FER over other methods, the baseline metrics reported by \cite{ref:16} are again used. Here, a number of state-of-of-the-art non-continual learning methods are used to evaluate the 21 emotion labels of the CFEE \cite{ref:17} database. The results are compared with our method as evaluated after the final continual learning iteration with 21 labels, and are displayed in Table \ref{tab:5}.

\begin{table}[!h]
\centering
\begin{tabular}{|l|l|}
\hline
\textbf{Method} & \textbf{Accuracy} \\
\hline
AlexNet \cite{ref:55} & 0.5637 \\
\hline
VGG-16 \cite{ref:56} & 0.5018 \\
\hline
VGG-19 \cite{ref:56} & 0.4971 \\
\hline
Inception-v3 \cite{ref:57} & 0.4288 \\
\hline
DenseNet-201 \cite{ref:58} & 0.4862 \\
\hline
SCN	\cite{ref:59} & 0.4621 \\
\hline
PSR	\cite{ref:60} & 0.5591 \\
\hline
ESRs \cite{ref:61} & 0.5781 \\
\hline
\textbf{Our Method} & \textbf{0.7176} \\
\hline
\textbf{Our Method (excl. singular labels*)} & \textbf{0.7182} \\
\hline
\end{tabular}
\caption{AveSA at final iteration of our method and non-continual learning methods}
\label{tab:5}
\end{table}

To evaluate the \textit{Few-shot Learning Phase}, the best performing test set from the \textit{Basic FER Phase} was used. The same hyperparameters were used as in the \textit{Continual Learning Phase}. One experiment was conducted for each of the 15 compound expression classes in the CFEE \cite{ref:17} dataset, making $k_{compound}=15$. Each of these 15 experiments were repeated in trials consisting of 5, 3 and 1 training samples. The test accuracy and number of steps to train the model to convergence (using early stopping) were recorded at the end of each experiment. The results are displayed in Table \ref{tab:6}.

\begin{table}[t]
\centering
\begin{tabular}{|l|l|l|l|l|l|l|}
\hline
\textbf{Expression} & \multicolumn{2}{l|}{\textbf{5-shot}} & \multicolumn{2}{l|}{\textbf{3-shot}} & \multicolumn{2}{l|}{\textbf{1-shot}} \\
\hline
& \textbf{Acc.} & \textbf{Steps} & \textbf{Acc.} & \textbf{Steps} & \textbf{Acc.} & \textbf{Steps} \\
\hline
Happily surprised & 1.0 & 3840 & 1.0 & 4800 & 1.0 & 5536 \\
\hline
Happily disgusted & 1.0 & 800 & 1.0 & 960 & 1.0 & 1344 \\
\hline
Sadly angry	& 1.0 &	480 & 1.0 & 384 & 1.0 & 608 \\
\hline
Angrily disgusted & 1.0 & 640 & 1.0 & 864 & 1.0 & 1056 \\
\hline
Appalled & 1.0 & 320 & 1.0 & 288 & 1.0 & 320 \\
\hline
Hatred & 1.0 & 640 & 1.0 & 672 & 1.0 & 896 \\
\hline
Angrily surprised & 1.0 & 640 & 1.0 & 768 & 1.0 & 1536 \\
\hline
Sadly surprised & 1.0 & 640 & 1.0 & 576 & 1.0 & 768 \\
\hline
Disgustedly surprised & 1.0 & 800 & 1.0 & 960 & 1.0 & 1504 \\
\hline
Fearfully surprised & 1.0 & 480 & 1.0 & 384 & 1.0 & 736 \\
\hline
Awed & 1.0 & 800 & 1.0 & 1056 & 1.0 & 1184 \\
\hline
Sadly fearful & 1.0 & 480 & 1.0 & 480 & 1.0 & 480 \\
\hline
Fearfully disgusted & 1.0 & 640 & 1.0 & 768 & 1.0 & 928 \\
\hline
Fearfully angry & 1.0 & 480 & 1.0 & 384 & 1.0 & 448 \\
\hline
Sadly disgusted	& 1.0 & 1280 & 1.0 & 1248 & 1.0 & 1472 \\
\hline
\end{tabular}
\caption{Accuracy and training time for few-shot learning experiments}
\label{tab:6}
\end{table}

\section{Conclusions}
\label{sec:6}

A novel method for compound facial expression recognition was developed by distilling the knowledge of basic expressions when learning new compound expressions. Two main experiments were conducted in continual learning and few-shot learning.

We demonstrate improvements in continual learning for complex FER through our novel knowledge distillation and Predictive Sorting Memory Replay techniques, achieving the state-of-the-art with 74.28\% Overall Accuracy on new classes only (an improvement of 0.67\%). The Overall Accuracy on all classes is 73.27\% which is comparable to baseline results and indicates a reduction in the effects of catastrophic forgetting as the accuracy on known classes is not greatly impacted when learning new classes. Results could potentially be further improved by tuning the \textit{Temperature} ($T$) and \textit{Distillation Weight} ($\gamma$) hyperparameters.

Our method also demonstrates an improvement in accuracy over other state-of-the-art non-continual learning methods for facial expression recognition by 13.95\%. This demonstrates the benefits of our approach to learning facial expressions through continual learning, by first learning to recognise basic facial expressions and then synthesising that knowledge to learn new complex facial expressions in a similar way to humans.

Our method achieves 100\% accuracy in all classes using only 5, 3 or 1 samples in few-shot learning, which is the state-of-the-art in the facial expression recognition domain to the best of our knowledge. These results also demonstrate the benefits of learning to recognise basic facial expressions prior to learning complex facial expressions, as the model was able to very quickly learn the new expression classes using only a limited number of image samples through knowledge distillation of basic features.

By visually inspecting the Grad-CAM heatmaps of basic expression images with those of compound expressions, a strong correlation was found between the activations of features in basic expressions and that of features in compound expressions. The activated areas also appear to align with the facial action encoding system \cite{ref:3}. In future works, the use of action units as an additional feature extraction component may also be able to increase the accuracy of the model for both basic facial expression recognition and complex facial expression recognition.

One limitation of this work is in evaluating the model using only one dataset. By evaluating on multiple datasets, we could get a better picture of the model's generalisation performance with a wider variety of subjects and image conditions. Evaluating the model on in-the-wild complex emotion datasets such as \textit{EmotioNet} \cite{ref:43} and \textit{AffectNet} \cite{ref:44} would also prove useful in assessing the model's generalisation and suitability for practical applications such as human-computer interaction, as these datasets more closely resemble real facial expressions.


\bibliographystyle{IEEEtran}
\bibliography{references.bib}

\begin{IEEEbiography}[{\includegraphics[width=1in,height=1.25in,clip,keepaspectratio]{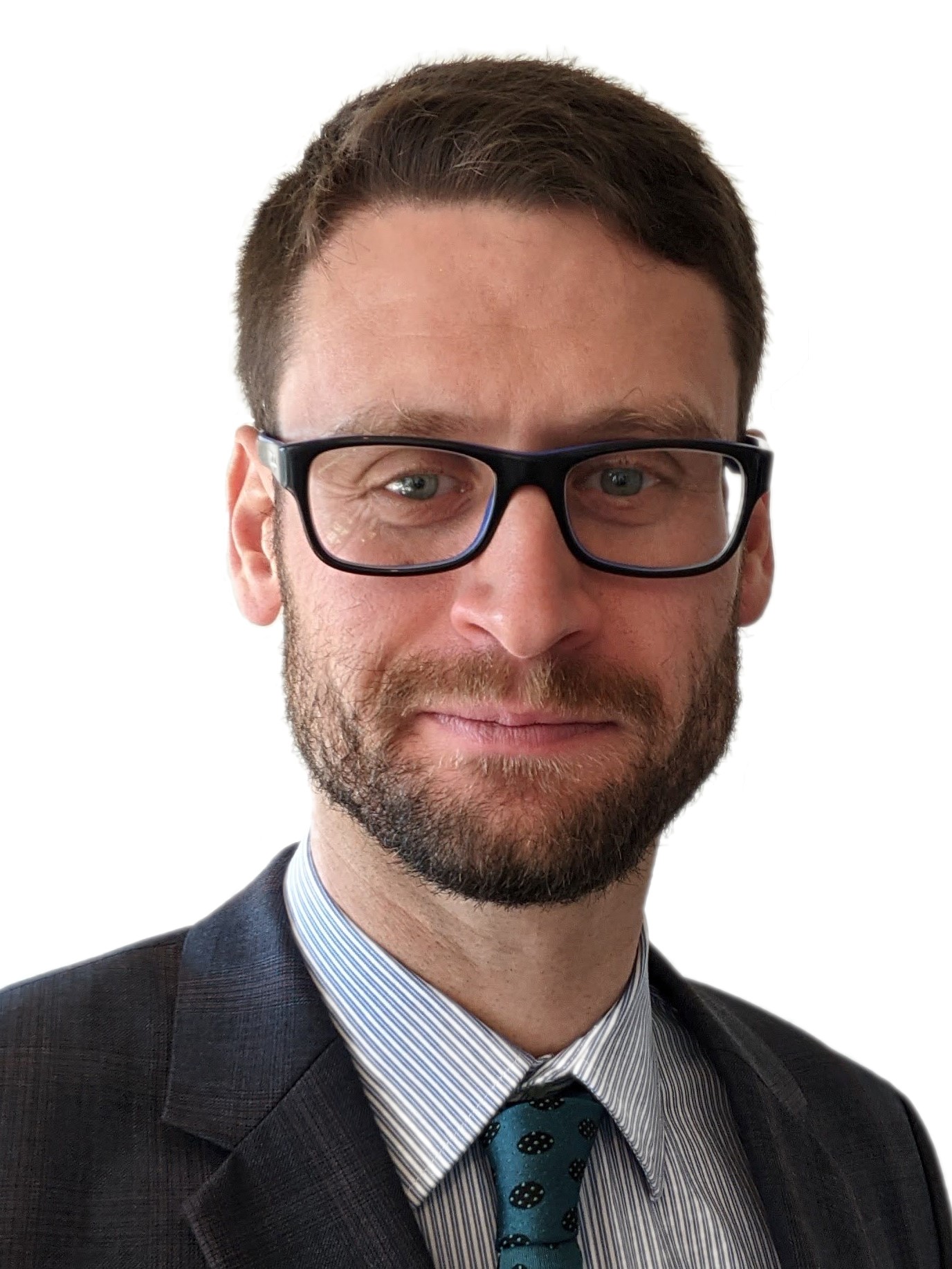}}]{Angus Maiden}{\space} completed the Master of Applied Artificial Intelligence (Professional) from Deakin University in July 2022, with a Weighted Average Mark of 91.7\% and a GPA of 4.0/4, for which he received the Deakin Scholarship for Excellence (Postgraduate), the Student Excellence Award (Research) for best paper, and the Student Excellence Award (Postgraduate) for achieving the highest overall weighted average mark in his cohort. He is also a member of the Golden Key International Honour Society.

Angus has been appointed as a Visiting Researcher at Deakin University since October 2022. His research interests include artificial intelligence, deep learning, machine learning, natural language processing, computer vision and emotion recognition. He is currently employed as a Data Scientist Technical Lead at a startup software company where he builds innovative health-tech products using machine learning and artificial intelligence.
\end{IEEEbiography}

\begin{IEEEbiography}[{\includegraphics[width=1in,height=1.25in,clip,keepaspectratio]{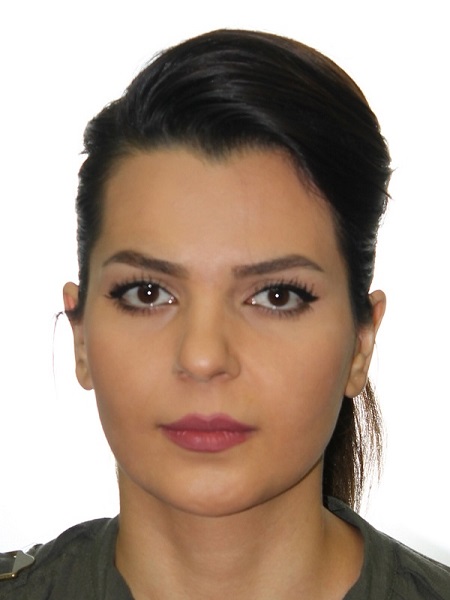}}]{Bahareh Nakisa}{\space} (Member, IEEE) received the Bachelor of Science degree in Software Engineering from Iran in 2008, the Master of Computer Science degree from the National University of Malaysia in 2014, and the Ph.D. degree in Computer Science (Artificial Intelligence) from the Queensland University of Technology (QUT), Australia in 2019.

Bahareh started working in the industry as an AI Scientist and the Lead AI Scientist. She then joined the School of Information Technology, Deakin University, as a Lecturer of applied AI, in 2019. She is currently a Lecturer of applied AI and the Course Director of applied AI at the School of Information Technology, Deakin University.

Bahareh's research interests include artificial intelligence, machine learning, and deep learning. She has applied analytic and algorithmic tools from these fields to solve real-world problems related to diverse domains, especially in health and affective computing. 
\end{IEEEbiography}

\end{document}